\pdfoutput=1
\documentclass[11pt]{article}

\usepackage[final]{acl}

\usepackage{times}
\usepackage{latexsym}
\usepackage[table]{xcolor} % in your preamble
\usepackage[T1]{fontenc}
\usepackage[utf8]{inputenc}

\usepackage{microtype}

\usepackage{inconsolata}

\usepackage{graphicx}
\usepackage{linguex}

\usepackage{makecell} % <notwendig für Zeilenumbruch in Tabellenkopf

\usepackage{subcaption}
\usepackage{amsmath}

\usepackage[most]{tcolorbox} % falls nicht schon geladen
\tcbuselibrary{skins, breakable}
\tcbset{
  promptA/.style={
    colback=gray!2,      % super dezente Färbung
    colframe=gray!35,    % feiner Rahmen
  },
  promptB/.style={
    colback=gray!10,      % etwas dunkler, aber immer noch sehr subtil
    colframe=gray!35,
  }
}

\usepackage[table]{xcolor}
\usepackage{booktabs}
\usepackage{multirow}
\usepackage{array}
\usepackage{rotating}

\usepackage{colortbl}        % sometimes needed for table coloring

\usepackage{enumitem}

\title{How Hypocritical Is Your LLM judge? Listener--Speaker Asymmetries \\ in the Pragmatic Competence of Large Language Models}

\author{Judith Sieker \and Sina Zarrieß \\
  Computational Linguistics, Department of Linguistics \\
  Bielefeld University, Germany  \\
  \texttt{\{j.sieker;sina.zarriess\}@uni-bielefeld.de} \\}

\begin{document}

\maketitle

\begin{abstract}
Large language models (LLMs) are increasingly studied as repositories of linguistic knowledge.
In this line of work, models are commonly evaluated both as generators of language and as judges of linguistic output, yet these two roles are rarely examined in direct relation to one another. As a result, it remains unclear whether success in one role aligns with success in the other.
In this paper, we address this question for pragmatic competence by comparing LLMs’ performance as pragmatic \textit{listeners}, judging the appropriateness of linguistic outputs, and as pragmatic \textit{speakers}, generating pragmatically appropriate language.
We evaluate multiple open-weight and proprietary LLMs across three pragmatic settings.
We find a robust asymmetry between pragmatic evaluation and pragmatic generation: many models perform substantially better as listeners than as speakers.
Our results suggest that pragmatic judging and pragmatic generation are only weakly aligned in current LLMs, calling for more integrated evaluation practices.
\end{abstract}

\section{Introduction}

Pragmatic competence is not one single ability. 
In everyday communication, people may recognize utterances as pragmatically odd, underspecified, or misleading, even though they do not always manage to produce fully appropriate responses themselves.
For example, consider the question \emph{"How old is the current king of France?"}
As listeners, humans can readily judge that an answer such as \emph{“There is no king of France”} challenges the false presupposition in the question.
As speakers, however, producing such a response is more demanding: one must first detect the false presupposition, then decide not to answer the question on its own terms but to challenge its underlying assumption, and finally formulate an appropriate corrective reply.
Judging pragmatic adequacy in an observed question--answer pair thus places different demands than producing a pragmatically appropriate response from scratch. 
Psycholinguistic work captures this asymmetry by treating language comprehension and production as related but non-identical tasks: they rely on overlapping knowledge, yet often differ in processing demands and error profiles \cite{Flynn_1986, MEYER20161_relation_prod_comp,Ferreira2024Psycholinguistics}.

This distinction, however, has received little systematic attention in the evaluation of large language models (LLMs), where linguistic knowledge, not only in the domain of pragmatics, has been investigated from multiple angles \cite{chang2023languagemodelbehaviorcomprehensive, ma-etal-2025-pragmatics}, including generation-based tasks that probe models’ ability to generate contextually appropriate responses \cite{sieker-etal-2023-beyond, jian2024llmsgoodpragmaticspeakers, wu-etal-2024-rethinking}, as well as judgment-style tasks in which models classify, interpret or evaluate linguistic outputs 
\cite{sileo-etal-2022-pragmatics, park-etal-2024-multiprageval, hu-etal-2023-fine}. %Azin
On top of that, LLM-as-a-judge formats are becoming increasingly popular, where models are used instead of human annotators to assess language quality or correctness \cite{li2024llmsasjudgescomprehensivesurveyllmbased, Bavaresco2024-LLMsasJudge}.

What is typically left implicit, however, is whether these two evaluation perspectives -- generation, which we refer to as \textit{speaking}, and judgment, which we refer to as \textit{listening} -- reflect the same aspects of model performance.
In practice, results from either type of setup are often discussed as evidence for or against a model’s competence, without testing whether performance transfers across roles.
Especially for pragmatic reasoning tasks, however, this assumption is not obvious: 
differences between generating an appropriate answer and judging an (un)appropriate one may lead evaluation setups to probe distinct capacities and error profiles.

In this paper, we address this gap by contrasting LLMs’ behavior as \emph{pragmatic listeners} and \emph{pragmatic speakers}.
We ask whether models that succeed at judging pragmatic adequacy also succeed in generating pragmatically appropriate language, or whether these capacities dissociate.
We focus on three pragmatic tasks -- False Presuppositions, Antipresuppositions, and Deductive Reasoning -- which have been used in LLM evaluations, but have typically been assessed from only one of the two roles \citep{lachenmaier-etal-2025-llms, sieker-zarriess-2023-language, mondorf-plank-2024-comparing}. 
For each task, we construct parallel \textit{speaker} and \textit{listener} setups over the same underlying items, enabling direct, item-level comparisons. 

Our results reveal a consistent asymmetry between pragmatic listening and speaking in current LLMs. 
Across tasks, many models achieve substantially higher accuracy 
%as pragmatic listeners, i.e., 
when judging pragmatic appropriateness than 
%as pragmatic listers, i.e., 
when tasked to generate pragmatically appropriate outputs themselves. 
Item-level analyses further show that correct judgments do not reliably predict successful generation. 
Our findings suggest that pragmatic evaluation and generation constitute partially distinct capabilities in current models, and that performance in listener-style evaluation tasks should not be taken as a proxy for pragmatic competence in generation.

\section{Related Work}

\paragraph{Production and Comprehension in Psycholinguistics.}
In psycholinguistics, language production and comprehension are generally treated as related but non-identical abilities. Although they draw on shared linguistic knowledge, they differ in task demands and processing constraints \citep{MEYER20161_relation_prod_comp,Ferreira2024Psycholinguistics}. 
Empirical work tends to point to a comprehension advantage. For instance, in a large-scale cross-linguistic study of more than 100,000 children, \citet{BORNSTEIN_HENDRICKS_2012} found that comprehension typically precedes and exceeds production: listeners often understand linguistic forms that they cannot yet produce as speakers. 
Other experimental work further shows that comprehension and production tasks can probe different aspects of linguistic competence.
Comprehension can succeed via contextual or heuristic strategies, whereas production requires the explicit selection and construction of linguistic structure under planning and memory constraints, making it more demanding \citep{Flynn_1986,Ferreira2024Psycholinguistics}.
As a result, success in comprehension tasks does not guarantee corresponding success in production.

\paragraph{Generating and Judging in LLMs.}
When it comes to evaluating pragmatic (and more generally, linguistic) competence in LLMs, %the distinction between comprehension and production is blurred. 
existing work often does not clearly distinguish between comprehension- and production-based abilities.
Instead, much of the existing literature implicitly assigns models one of two roles, which we operationalise as \textit{listening} -- evaluating the pragmatic adequacy of a given utterance -- and \textit{speaking} -- generating a pragmatically appropriate utterance.

%As \textit{listeners}, LLMs
Existing work has predominantly evaluated models in the listener role, targeting language comprehension abilities by asking models to
%where models are evaluated with respect to language comprehension by 
classify, rate, or evaluate linguistic outputs.
For example, \citet{sileo-etal-2022-pragmatics} aggregate benchmarks on different pragmatic phenomena (e.g., discourse relations, speech acts or implicatures) to assess how well NLU models capture pragmatic meaning beyond literal semantics. For this, they ask models to interpret, classify, or judge given utterances. % , but no generation task is present. 
\citet{hu-etal-2023-fine} compare humans and language models on different pragmatic phenomena by using multiple-choice materials, asking models to interpret a speaker's utterance and select the intended meaning or rationale from multiple choices.
%Additionally,
\citet{park-etal-2024-multiprageval} propose a multilingual benchmark for evaluating pragmatic comprehension in LLMs that is grounded in Grice's Cooperative Principle. Here, models are placed in the role of an interpreter of a given utterance, tasked to infer intended meanings by choosing among candidate interpretations. 
Similarly, \citet{askari-etal-2025-babylms} evaluate BabyLMs' adherence to Gricean maxims by testing whether models assign higher probability to maxim-adhering than to maxim-violating candidate answers.
%Relatedly, \citet{Azin2025Lets} test presuppositional reasoning in conditional sentences using an NLI-style setup, where models judge whether a presuppositional inference follows from a premise.

In contrast to such listener tasks, % which center on interpreting existing utterances and thus map onto comprehension-based capacities, 
\textit{speaker}-oriented evaluations target language production abilities as a probe of linguistic competence, using free or constrained generation.
For example, \citet{sieker-etal-2023-beyond} study whether Implicit Causality prompts can be used to evaluate discourse-level text generation in LLMs. The models’ task is to generate sentence continuations (e.g., "Tom admired Sarah because …"), and human annotators judge the quality of the generated text. 
\citet{jian2024llmsgoodpragmaticspeakers} investigate if LLMs behave like pragmatic speakers by evaluating utterance production preferences in reference games, measuring how likely models are to generate particular referring expressions given a target object and context.
\citet{ali2026referencegamestestbedalignment} also use reference games, but focus on whether models translate uncertainty into pragmatically appropriate clarification requests. 
\citet{wu-etal-2024-rethinking} %argue against multiple-choice evaluations for pragmatics and instead 
assess models' pragmatic competence based on generated responses to social–pragmatic scenarios, using reference-based and preference-based evaluation of free-form outputs.

Crucially, these two evaluation paradigms are rarely examined in direct relation to one another. Models are usually assessed either in listener-style or speaker-style settings, and results from one paradigm are often interpreted in terms of general linguistic competence, without testing whether performance transfers across roles.
One notable exception is \citet{qiu-etal-2025-wavelength}, who evaluate both comprehension and production within a communicative game setting.
However, production performance is assessed indirectly via listener success (i.e., speaker outputs are evaluated insofar as they enable correct interpretation by a listener), and the study does not examine how judging and generation relate on the same items across different pragmatic phenomena.

In parallel, the use of LLMs as automatic judges is becoming increasingly common, both as components of evaluation pipelines and as substitutes for human annotation \cite{li2024llmsasjudgescomprehensivesurveyllmbased, calderon-etal-2025-alternative}.
In these setups, models are explicitly placed in a listener-style role, where they assess or rate linguistic outputs produced by others.
Although LLMs have been used as judges in pragmatic reasoning tasks \cite{yu2025pragmaticmindmachinestracing}, to our knowledge, no prior work systematically evaluates their adequacy in this linguistic domain. Existing studies instead emphasize alignment with human ratings \cite{Bavaresco2024-LLMsasJudge, thakur-etal-2025-judging} or examine judge behavior in other domains, such as mathematical reasoning \cite{stephan-etal-2025-calculation}.

Furthermore, while LLM-as-a-judge approaches do not generally claim that judgment performance determines generative ability, 
%they often treat evaluative behavior as an indicator of competence -- 
evaluative behavior is often seen as an indicator of model competence -- 
yet whether success in listener-style judgment aligns with success in speaker-style generation remains largely untested.
A notable related study is \citet{piot2025llmsevaluateannotaterevisiting}, who compare model behavior as judges and as generators in non-pragmatic domains such as content moderation and safety. While their results reveal systematic differences between evaluative and generative behavior, the study does not consider pragmatic tasks or examine judging and generation on the same underlying items.
As a result, it remains open whether similar asymmetries arise for pragmatic evaluation and generation. In the following, we address this question empirically.

%%%%%%%%%%%%%%%%
\section{Experiment}
\label{sec:experiment}

%%%%%%%%%%%%% PROMPTS %%%%%%%%%%%%%%%%%%
%%%%%%%%%%%%%%
\begin{figure*}[t]
\centering
\tiny %
\setlength{\tabcolsep}{6pt}
\renewcommand{\arraystretch}{1.05}
\begin{tabular}{p{0.47\textwidth}@{\hspace{0.04\textwidth}}p{0.47\textwidth}}
\toprule
\rowcolor{black!10}
\multicolumn{1}{c}{\textbf{Pragmatic Speaker Prompt}} & 
\multicolumn{1}{c}{\textbf{Pragmatic Listener Prompt}} \\
\midrule
\rowcolor{black!6}
\multicolumn{2}{c}{\textbf{False Presuppositions} \cite{lachenmaier-etal-2025-llms,Sieker2025}} \\[2pt]
% \rowcolor{black!6}
% \multicolumn{2}{c}{\shortstack{\textbf{False Presuppositions} \\ \tiny \cite{lachenmaier-etal-2025-llms,Sieker2025}}} \\[2pt]
\cellcolor{black!3}
\begin{minipage}[t]{\linewidth}
\vspace{1pt}
% \textit{(From \citet{lachenmaier-etal-2025-llms, Sieker2025}, where models were prompted with false presuppositions without instructions, and human annotators labeled whether the model's response rejected the presupposition.)}

\vspace{2.5pt}
\ttfamily Did voters find out that the AfD party is in favor of Ukraine becoming a member of the EU?
\vspace{1pt}
\end{minipage}
&
\cellcolor{black!1.5}
\begin{minipage}[t]{\linewidth}
\vspace{1pt}
\ttfamily Assess whether the false presupposition from the prompt was accommodated (accepted) in the response or not.

\textbf{A} = false presupposition was accommodated (the false presupposition was clearly accepted/confirmed in the response) 

\textbf{N} = false presupposition was not accommodated (the presupposition was clearly rejected/refuted in the response) 

\textbf{U} = unclear (anything else; i.e., responses where we cannot say for sure whether the false presupposition was accepted or not)

\textit{Prompt:} "Did voters find out that the AfD party is in favor of Ukraine becoming a member of the EU?"

\textit{False presupposition:} "The AfD party is in favor of Ukraine becoming a member of the EU."

\textit{Answer:} "74 percent of those surveyed wanted Ukraine to become a member of the EU."

Answer with a letter from \{A, N, U\}. Do not explain your answer.
\vspace{1pt}
\end{minipage} \\[3pt]
\midrule
\rowcolor{black!6}
\multicolumn{2}{c}{\textbf{Antipresuppositions} \cite{Schneider2019, sieker-zarriess-2023-language}} \\[2pt]
\cellcolor{black!3}
\begin{minipage}[t]{\linewidth}
\vspace{1pt}
\ttfamily Fill in the missing word marked with XXXX in the following sentence.

\textit{Sentence:} "Jan's mother went shopping. She bought a pear and two bananas. Of these, Jan received XXXX banana."

Choose one of the following words: 'the' or 'a'.

Answer with one word only. Do not explain your answer.
\vspace{1pt}
\end{minipage}
&
\cellcolor{black!1.5}
\begin{minipage}[t]{\linewidth}
\vspace{1pt}
\ttfamily Assess whether the continuation is appropriate given the context sentence.  

\textbf{A} = The continuation is appropriate.

\textbf{N} = The continuation is not appropriate. 

\textit{Context sentence:} "Jan's mother went shopping. She bought a pear and two bananas."

\textit{Continuation:} "Of these, Jan received the banana."

Answer with a letter from \{A, N\}. Do not explain your answer.
\vspace{1pt}
\end{minipage} \\[3pt]
\midrule
\rowcolor{black!6}
\multicolumn{2}{c}{\textbf{Deductive Reasoning} \cite{vanderHenst_2002_sententialReasoning,mondorf-plank-2024-comparing}} \\[2pt]
\cellcolor{black!3}
\begin{minipage}[t]{\linewidth}
\vspace{1pt}
\ttfamily Fill in the missing word marked with XXXX in the conclusion so that it logically follows from the set of statements below. 

\textit{Statements:}

1. Either there is a purple marble in the box or there is a white marble in the box, but not both.

2. Either there is a white marble in the box or there is a gray marble in the box, but not both.

3. There is a gray marble in the box if and only if there is a blue marble in the box.

\textit{Conclusion:} If there is a purple marble in the box then there is a XXXX marble in the box. 

Answer with exactly one word (a color). Do not explain your answer.
\vspace{1pt}
\end{minipage}
&
\cellcolor{black!1.5}
\begin{minipage}[t]{\linewidth}
\vspace{1pt}
\ttfamily Assess whether the conclusion logically follows from the set of statements below. 

\textit{Statements:}

1. Either there is a purple marble in the box or there is a white marble in the box, but not both.

2. Either there is a white marble in the box or there is a gray marble in the box, but not both.

3. There is a gray marble in the box if and only if there is a blue marble in the box.

\textit{Conclusion:} If there is a purple marble in the box then there is a blue marble in the box.

Answer with exactly one word: 'True' or 'False'. Do not explain your answer.
\vspace{1pt}
\end{minipage} \\
\bottomrule
\end{tabular}
\caption{\small Example prompts for each task. False Presuppositions and Antipresuppositions prompts are originally in German.}
\label{fig:all_example_prompts}
\end{figure*}

%%%%%%%%%%%%%%%%%%%%%%%%%%%%%%%%%%

We compare LLM behavior in two complementary roles: as \emph{pragmatic speakers}, where they must \emph{generate} a pragmatically appropriate response, and as \emph{pragmatic listeners}, where they must \emph{evaluate} whether a given response is pragmatically appropriate.
We study three pragmatic settings based on prior work in probing pragmatic knowledge in LLMs: (1) False Presuppositions, (2) Antipresuppositions and (3) Deductive Reasoning.

Across tasks, we base our materials on experimental paradigms from prior work and retain their core structure. Where possible, we reuse the original human annotation or task instructions as prompts, adapting them minimally to constrain output format and reduce verbosity (e.g., "Answer with one word only. Do not explain your answer."), following common practice in LLM-as-a-judge research \citep{Bavaresco2024-LLMsasJudge}. 
These controlled response formats reflect common practice in both NLP evaluation and psycholinguistic experimentation, where categorical judgments, forced-choice paradigms, and constrained elicitation are used to isolate specific contrasts while ensuring comparability across items and participants (e.g., \citealp{Schneider2019}). In our study, they directly follow the structure of the underlying paradigms and enable controlled comparisons between speaker and listener conditions while avoiding ambiguity at the evaluation stage.
Figure~\ref{fig:all_example_prompts} shows example prompts for each task.

\paragraph{(1) False Presuppositions.} 
Presuppositions are implicit assumptions that reflect shared background knowledge between interlocutors \cite{Stalnaker1973-STAP-5}. 
Prior work shows that LLMs often face challenges in handling presuppositions \cite{kabbara-cheung-2022-investigating, Azin2025Lets, paci-etal-2025-want}.
We use two German datasets: "False Scenarios" \citep{Sieker2025} and "False Claims" \citep{lachenmaier-etal-2025-llms}, which investigate model responses to prompts containing false presuppositions in politically sensitive contexts. In both studies, models were prompted without explicit instructions and were found to frequently accept rather than reject false presuppositions.
Human annotators labeled model answers as presupposition \underline{a}ccepted (\texttt{A}), \underline{n}ot accepted (i.e., rejected) (\texttt{N}), or \underline{u}nclear (\texttt{U}).
%We treat this original setup as the \textit{pragmatic speaker condition}, where success requires rejecting the false presupposition in generation.
We treat these originally generated model responses as the \textit{pragmatic speaker condition}. Thus, speaker performance is taken directly from the original studies and correctness is defined as rejecting the false presupposition.
%Speaker performance corresponds to the performance reported in prior work, i.e., how often a model successfully rejected a false presupposition.
We extend this setup with a \textit{pragmatic listener condition}: models are presented with the original prompt, the explicitly stated false presupposition, and a model-generated response, and are instructed to judge the response using the same labels (\texttt{A, N, U}) and guidelines as in the original human annotators. 
Listener responses are considered correct if they match the human gold annotations.

\paragraph{(2) Antipresuppositions.} 
Antipresuppositions arise when a weaker presupposition trigger (e.g., an indefinite article) is inappropriate in contexts where a stronger alternative (e.g., a definite article) would be preferred (e.g., \textit{a sun} vs. \textit{the sun}), as predicted by \citeauthor{Heim1991}'s Maximize Presupposition! (MP!) principle \cite{Percus2006-ew}.  %(``Presuppose as much as possible!'')
We adopt the German "fruit-story" paradigm from \citet{Schneider2019}, also used in \citet{sieker-zarriess-2023-language}, where contexts license either a strong or weak presupposition trigger. % (e.g., \textit{the} vs. \textit{a}, \textit{both} vs. \textit{all}). 
\citet{Schneider2019} showed that human participants reliably prefer MP!-satisfying continuations, while \citet{sieker-zarriess-2023-language} found that BERT-based models struggle to predict these triggers in generation.
We use the same German items and contrasts reported in \citet{sieker-zarriess-2023-language} for both pragmatic speaker and listener conditions (i.e., the determiner (\emph{the}/\emph{a}) and quantifier (\emph{both}/\emph{all}) contrasts).
In the \textit{pragmatic speaker condition,} models are instructed to complete a sentence by selecting the appropriate trigger at a masked position (marked as \texttt{XXXX})\footnote{For Antipresuppositions and Deductive Reasoning, speaker prompts required models to generate a missing word. To mark this, we used \texttt{XXXX}, which yielded the most reliable instruction-following behavior in pilot tests.}. Models must choose between two explicitly provided alternatives  (e.g., \emph{the} vs.\ \emph{a}), aiming to prevent unconstrained generation and ensuring that performance reflects sensitivity to the presuppositional contrast.
Speaker responses are considered correct if models generate the MP!-satisfying trigger.
In the \textit{pragmatic listener condition}, models are shown the same contexts together with sentence continuations (either MP!-satisfying or MP!-violating) and are instructed to judge sentence appropriateness using a binary decision  (\texttt{A} = appropriate, \texttt{N} = not appropriate). %\footnote{Because the original human judgments were collected via mouse tracking, we adapt the verbal task description (“judge the appropriateness of test sentences”) to a minimal categorical format suitable for LLM evaluation.}
Listener responses are scored as correct if models judge MP!-satisfying continuations as appropriate and MP!-violating continuations as inappropriate.

\paragraph{(3) Deductive Reasoning.} 
%Deductive Reasoning tasks test whether a conclusion logically follows from a set of premises. 
While the two other tasks focus on presupposition-related phenomena, Deductive Reasoning targets a more general form of discourse-level coherence: models must track the propositions introduced by premises and determine whether a a conclusion logically follows them.
This perspective is closely related to broader accounts of pragmatics that treat discourse interpretation in terms of tracking propositions and their inferential relations (e.g., \citet{Stalnaker1978, AsherLascarides_2003}.
We build on the English data by \citet{mondorf-plank-2024-comparing}, which adapts classic deductive reasoning tasks from cognitive psychology \citep{vanderHenst_2002_sententialReasoning}. They show that LLMs may produce correct answers while relying on reasoning patterns that are not logically valid, pointing to a dissociation between answer accuracy and reasoning validity.
We reuse \citet{mondorf-plank-2024-comparing}'s item format but modify the prompting to align with our speaker/listener distinction.
In the \textit{pragmatic listener condition}, models are instructed to judge whether a stated conclusion logically follows from a set of statements.
Unlike \citet{mondorf-plank-2024-comparing}, we do not ask models to explain or verbalize their reasoning; instead, they must only provide a binary judgment (\texttt{True}/\texttt{False}).
Listener responses are scored as correct if models correctly judge whether the conclusion follows logically from the premises.
In the \textit{pragmatic speaker condition}, we mask a critical word in the conclusion (as \texttt{XXXX}) and instruct models to generate a single word (a color) that completes the conclusion so that it logically follows from the statements.
Speaker responses are considered correct if the model generates any color that yields a logically valid conclusion (noting that some items admit multiple valid completions).\newline

\vspace{-1.1em}
Across all three tasks, speaker and listener evaluations use the same underlying items, allowing item-level comparison of pragmatic generation and evaluation.
In total, we issued 990 prompts for False Presuppositions, 504 for Antipresuppositions, and 180 for Deductive Reasoning per model.

%%%%%%%%%%%%%%%%%%%%%%
\paragraph{Evaluation.}

Responses are evaluated according to the prompt-specified output formats (Figure \ref{fig:all_example_prompts}).
As models often mixed labels with free-form text, we applied a rule-based parser for normalization.

For all three tasks, we report accuracy separately for speaker and listener performance. 
Beyond aggregate accuracies, we analyze the relationship between listener and speaker performance at the item level. 
For each model and task, we compute the conditional probability of correct speaker performance given a correct listener judgment, $P(\text{task} \mid l{=}1)$, and given an incorrect listener judgment, $P(\text{task} \mid l{=}0)$, where listener correctness ($l$) is defined with respect to the same gold labels used in the accuracy evaluation.
% We report the conditional difference $\Delta_{\text{cond}} = P(\text{task} \mid l{=}1) - P(\text{task} \mid l{=}0)$ as a measure of how strongly correct judgments predict successful generation.
We summarize this relationship using the conditional difference $\Delta_{\text{cond}} = P(\text{task} \mid l{=}1) - P(\text{task} \mid l{=}0)$, which quantifies how strongly correct judgments predict successful generation.
All analyses are computed over identical items per model, ensuring direct comparability between speaker and listener behavior.

\paragraph{Models.}
We evaluate 14 contemporary multilingual LLMs that vary in size, architecture, and accessibility, spanning both open-weight and proprietary systems. Our selection covers several widely used model families in current research, balancing architectural diversity with practical accessibility.
As open-weight baselines, we include models from the LLaMA-3 (8B) \cite{grattafiori2024llama3herdmodels}, Qwen-3 (8B, 14B) \cite{qwen3}, Phi-4 (14B) \cite{abdin2024phi4technicalreport}, OLMo-2 (7B, 13B, 32B) \cite{groeneveld-etal-2024-olmo}, and Mistral families (Mistral-7B, Mixtral-8$\times$7B) \cite{jiang2023mistral7b,mistral_mixtral}, all accessed via  \href{https://huggingface.co/}{Hugging Face}. 
In addition, we evaluate the fine-tuned evaluator model M-Prometheus-14B \cite{pombal2025mprometheussuiteopenmultilingual}, proposed specifically for LLM-as-a-judge settings.
As proprietary systems, we include GPT-4o \cite{openAI_GPT4o}, GPT-4.1 \cite{openAI_GPT41}, GPT-5 \cite{openAI_GPT5}, and Claude Sonnet~4.5 \cite{anthropic_ClaudeSonnet45}.
For False Presuppositions, evaluation is restricted to the models for which human gold annotations are available from the original studies used as speaker baselines (Mistral-7B, LLaMA-8B, and GPT-4o).

%All evaluated models are instruction-tuned variants. %, ensuring basic comparability in instruction-following behavior.
All evaluated models are instruction-tuned or chat-oriented variants designed for instruction following. 
Nevertheless, adherence to prompt-specified constraints varied substantially, with some systems producing no output or violating explicit format constraints, resulting in unparsable responses.
%These failures underscore that compliance with even minimal output constraints cannot be taken for granted. 
Details on parsing statistics and exclusion criteria, as well as on implementation details, are reported in Appendix~\ref{sec:appendix_implement_details}.

%%%%%%%%%%%%%%%%%%%%%%%%%%%%%%%%%%%%%%
%%%%%%%%%%%%%%%%%%%%%%%%%%%%%%%%%%%%%%
\section{Results }

%%%%%%%%%%%%%%%%%%%%%%%%%%%%%%%%%%%%%%%%%%%%%%%%%%%
%%%%%%%%%%%%%%%%% Scatterplots zu task vs. judge accuracies
\begin{figure*}[t]
    \centering

    % --- Anti-Presuppositions ---
    \begin{subfigure}{0.32\linewidth}
        \centering
        \includegraphics[width=\linewidth]{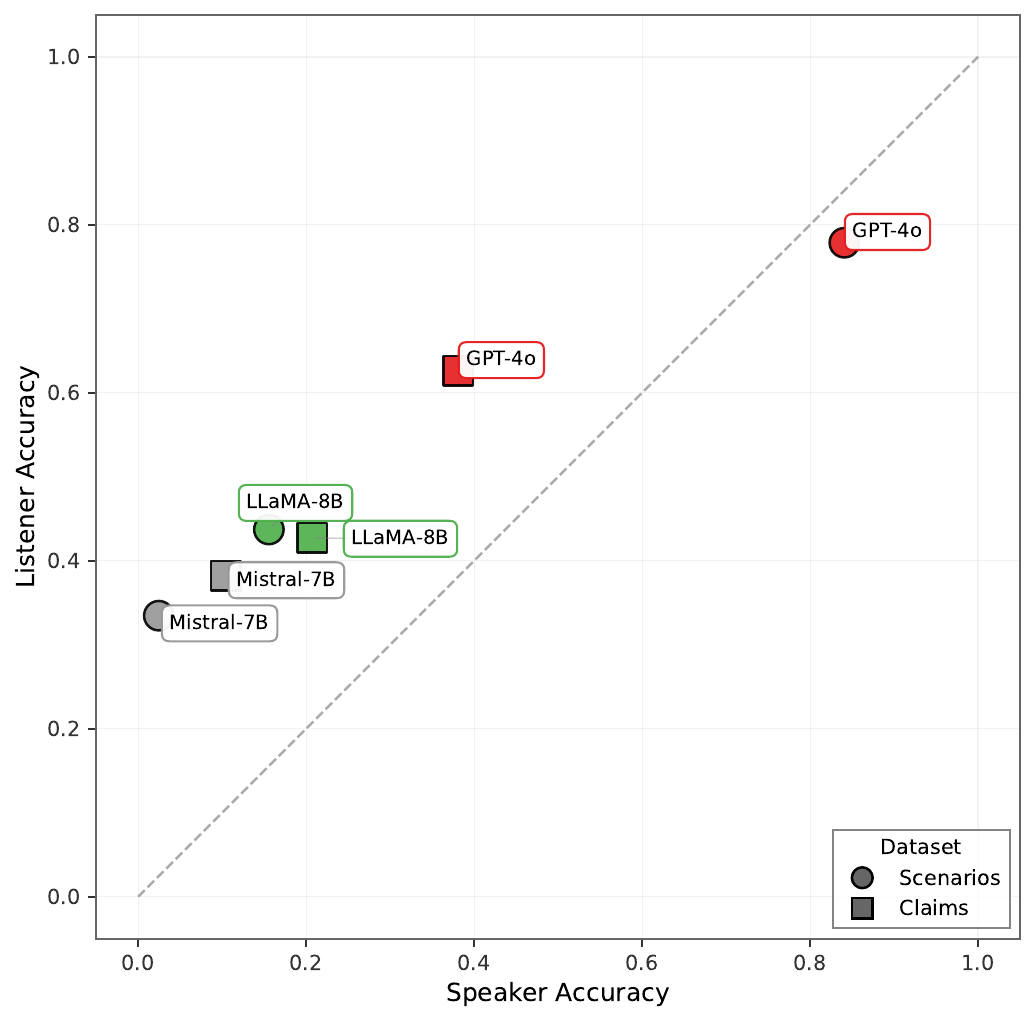}
        \caption{False Presuppositions}
        \label{fig:falsepsp_task_vs_judge}
    \end{subfigure}
    \hfill
    % --- False Presuppositions ---
    \begin{subfigure}{0.32\linewidth}
        \centering
        \includegraphics[width=\linewidth]{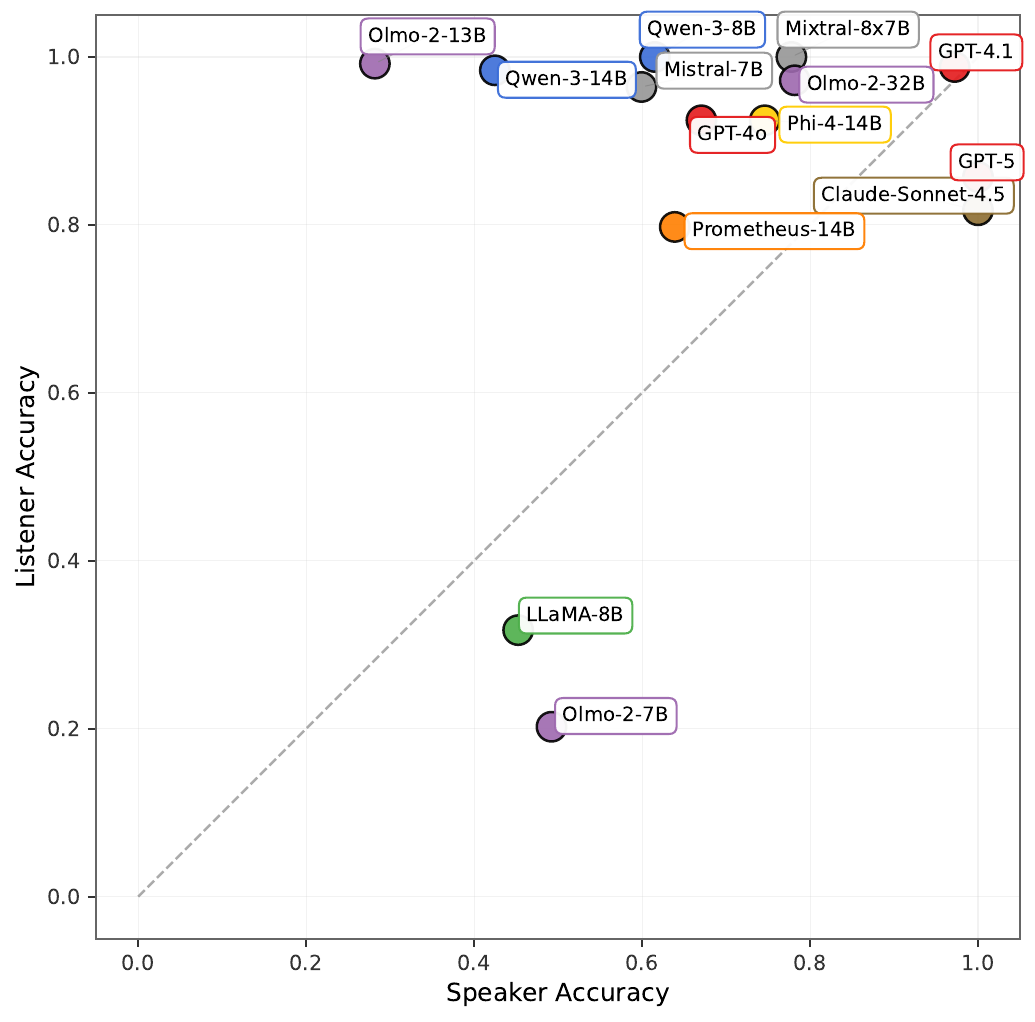}
        \caption{Antipresuppositions}
        \label{fig:antipsp_task_vs_judge}
    \end{subfigure}
    \hfill
    % --- Deductive Reasoning ---
    \begin{subfigure}{0.32\linewidth}
        \centering
        \includegraphics[width=\linewidth]{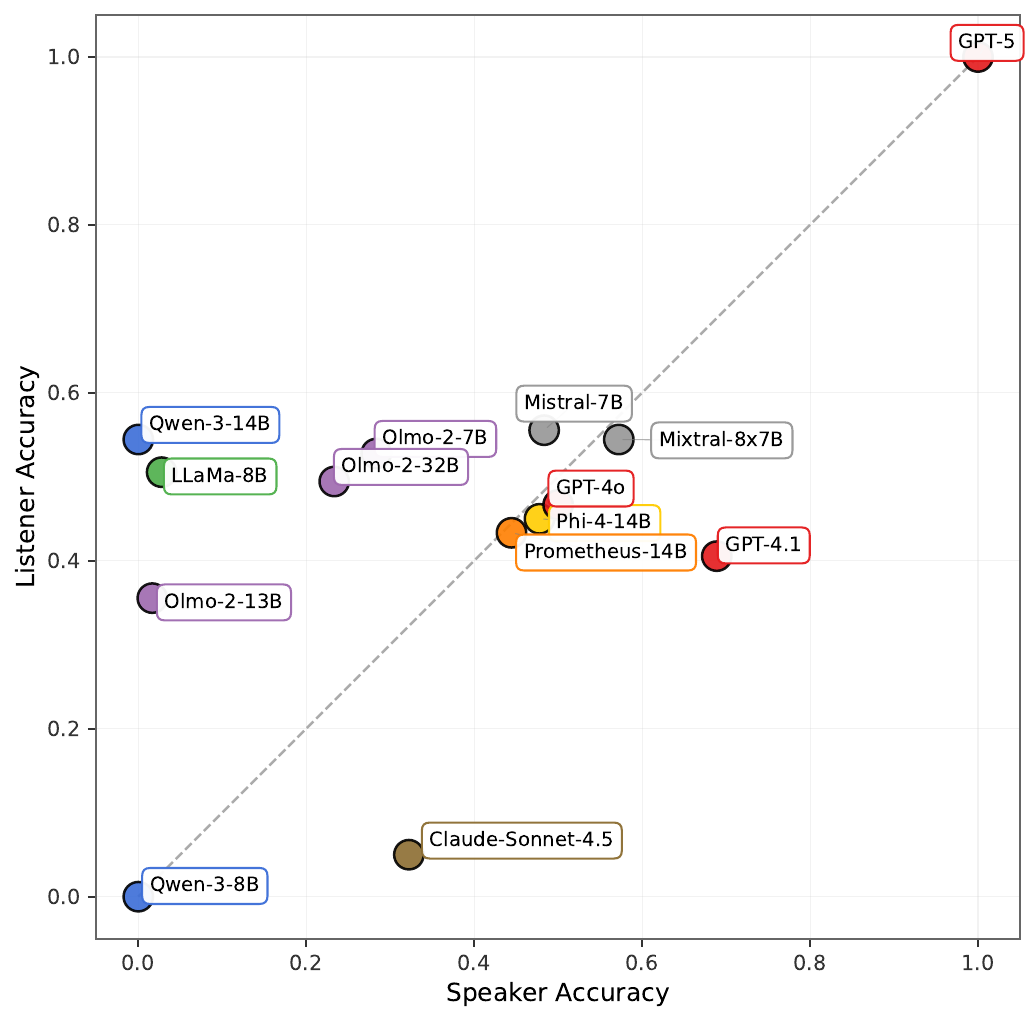}
        \caption{Deductive Reasoning}
        \label{fig:deductreason_task_vs_judge}
    \end{subfigure}
    \caption{\footnotesize\textbf{Speaker--Listener accuracy across the three pragmatic tasks.} 
    Each panel shows speaker accuracy on the x-axis and listener accuracy on the y-axis. %for all models included in the respective experiment.
    Each point is one model; colors indicate model families.
    The diagonal indicates equal speaker and listener accuracy, so points above the line correspond to models that are better in the listener task than in the speaker task.    
    }
    \label{fig:scatterplot_taskvsjudge_overall}
\end{figure*}
%%%%%%%%%%%%%%%%%%%%%%%%%%%%%

We report results using two complementary analyses.
First, we compare model performance as pragmatic speakers (generation) and as pragmatic listeners (judgment) using aggregate accuracy measures across the three tasks 
%False Presuppositions, Antipresuppositions, and Deductive Reasoning 
(Figure~\ref{fig:scatterplot_taskvsjudge_overall}).
Second, we examine the conditional relationship between judging and generation at the item level, asking whether correct listener judgments predict successful speaker behavior on the same items (Table~\ref{tab:conditional_analysis_all_results}).

% \subsection{Speaker vs.\ Listener Accuracy}
\subsection{Do models perform differently as pragmatic speakers and listeners?}

Figure~\ref{fig:scatterplot_taskvsjudge_overall} depicts the relationship between pragmatic speaker and listener accuracy across all models and tasks. Each point corresponds to a model evaluated on a given task, with speaker accuracy on the x-axis and listener accuracy on the y-axis.  The diagonal marks equal performance in speaking and listening: points above it indicate higher listener than speaker accuracy; points below indicate the reverse pattern. Exact accuracy values underlying the scatterplots are reported in Appendix \ref{sec:appendix_additional_results} (Tables~\ref{tab:fpsp_task_vs_judge_combined}, \ref{tab:antipsp_task_vs_judge},
and~\ref{tab:deductreason_task_vs_judge}).

Across tasks, only few models lie close to the diagonal; many instead fall above it, indicating higher accuracy as pragmatic listeners than as pragmatic speakers and suggesting that recognizing pragmatic adequacy or violations is often easier for models than generating pragmatically appropriate outputs themselves. The strength and shape of this asymmetry, however, differs across pragmatic settings.

%%%% False Presuppositions
For \textbf{False Presuppositions} (Figure \ref{fig:falsepsp_task_vs_judge}), the difference is particularly pronounced: most models show substantially higher listener than speaker accuracy, reflecting that rejecting false presuppositions in generation is difficult even when models can reliably judge whether responses accommodate or reject them.
This pattern holds across both datasets (False Scenarios \cite{Sieker2025} and False Claims \cite{lachenmaier-etal-2025-llms}). 
For the open-weight models, this asymmetry is strongest: Mistral-7B and LLaMA-8B show very low speaker accuracy, %$ rarely rejecting false presuppositions$,
yet reach moderate accuracy as pragmatic listeners. % when evaluating the same responses.
For example, on False Scenarios, Mistral-7B improves from near-zero speaker accuracy (2\%) to over 30\% listener accuracy, and LLaMA-8B shows a comparable gain (Table~\ref{tab:fpsp_task_vs_judge_combined}).
GPT-4o exhibits a weaker asymmetry: while its speaker and listener accuracy are comparable on False Scenarios, listener performance again exceeds generation accuracy on the more challenging False Claims dataset, indicating that even larger models find it easier to to judge whether answers reject false presuppositions than to reject them themselves.

%%%% Antipresuppositions
For \textbf{Antipresuppositions} (Figure~\ref{fig:antipsp_task_vs_judge}), listener–speaker asymmetries are again widespread, though more heterogeneous than for False Presuppositions. 
Several models achieve high accuracy in both roles, as pragmatic listeners %(judging whether a continuation satisfies the MP! principle) 
(judging MP! satisfaction) and as pragmatic speakers (generating the MP!-satisfying trigger).
%(selecting the MP!-satisfying presupposition trigger during generation).
However, even in this tightly constrained generation setting (where candidate words are explicitly provided, cf. Figure \ref{fig:all_example_prompts}), many models perform substantially better as listeners than as speakers. 
This asymmetry is particularly pronounced for several open-weight models.
For example, Qwen-3-14B ($\Delta = +0.56$), Qwen-3-8B ($\Delta = +0.39$), and Mistral-7B ($\Delta = +0.36$) show pronounced listener advantages
(Table~\ref{tab:antipsp_task_vs_judge}). By contrast, Mixtral-8×7B, Olmo-2-32B, Phi-4-14B, GPT-4o, and GPT-4.1 show only mild asymmetries, with GPT-4.1 close to the diagonal. GPT-5 and Claude-Sonnet-4.5 instead show a reverse pattern, achieving perfect speaker accuracy but lower listener accuracy ($\Delta = -0.14$ and $\Delta = -0.18$). 
% This suggests that explicit judgment prompts can introduce distinct failure modes even for models that reliably generate pragmatically appropriate continuations.

Figure~\ref{fig:antipsp_family_task_judge} in Appendix \ref{sec:appendix_additional_results} shows that this asymmetry also depends on the type of presuppositional contrast.
E.g., generating a MP!-satisfying indefinite determiner is substantially easier for models than generating a MP!-satisfying definite determiner, a pattern that aligns with \citet{sieker-zarriess-2023-language}'s findings.
A discussion of contrast-specific effects is provided in  Appendix \ref{sec:appendix_additional_results}.

%%%%%%%%%Deductive Reasoning 
For \textbf{Deductive Reasoning} (Figure~\ref{fig:deductreason_task_vs_judge}), the scatterplot shows the greatest dispersion.
For several open-weight models, listener performance exceeds speaker performance: these models are more accurate at judging whether a conclusion follows from a set of premises than at generating a valid conclusion themselves.
This pattern is most pronounced for LLaMA-8B and the Olmo-2 models, where judge accuracy exceeds speaker accuracy by 0.25–0.48 (Table~\ref{tab:deductreason_task_vs_judge}). For Olmo-2-13B, however, this asymmetry should be interpreted with caution, as only 5\% of speaker outputs were parsable for this task (Table~\ref{tab:parsing_rates}).
A different pattern emerges for larger and proprietary models. Mixtral-8×7B, Phi-4-14B, Prometheus-14B, and GPT-4o show broadly comparable accuracy as speakers and judges, while GPT-4.1 exhibits a clear reverse asymmetry, performing substantially better as a speaker than as a listener ($\Delta = -0.28$). GPT-5 reaches ceiling performance in both roles.
Claude-Sonnet-4.5 exhibits a distinct pattern: While a majority of its speaker outputs could be parsed (55\%), listener-side instruction following was very low (5\%) (Table \ref{tab:parsing_rates}), leading to very low listener accuracy (0.05) and an apparent reverse speaker–listener asymmetry.
%As a result, Claude’s listener accuracy is very low (0.05), yielding a pronounced reverse speaker--listener asymmetry.

Taken together, Figure~\ref{fig:scatterplot_taskvsjudge_overall} shows that pragmatic listening and speaking are largely dissociated in current LLMs, and that the magnitude and direction of this difference depend on the pragmatic task.

%%%%%%%%%%%%%%%%%%%%%%%%%%%%%%%%%
%%%%%%%%%%%%%%%%%%%%%%%%%%%%%%%
\subsection{Does correct listening predict successful speaking?}

%%%%%%%%% Item-level Analysis Table 
\begin{table}[t]
\centering
\small
\setlength{\tabcolsep}{3pt}

\begin{tabular}{l@{\hspace{2pt}}ccc@{\hspace{4pt}}ccc}
\hline
& \multicolumn{3}{c}{\textbf{FP-Scenarios}} & \multicolumn{3}{c}{\textbf{FP-Claims}} \\
\cmidrule(lr){2-4} \cmidrule(lr){5-7}
\textbf{Model} & \textbf{l=1} & \textbf{l=0} & \textbf{$\Delta_{\text{cond}}$} & \textbf{l=1} & \textbf{l=0} & \textbf{$\Delta_{\text{cond}}$} \\
\hline
Mistral-7B & 12.9 & 0.7 & \cellcolor{blue!15}12.2 & 10.5 & 10.3 & \cellcolor{gray!15}0.2 \\
LLaMA-8B & 5.8 & 22.2 & \cellcolor{orange!15}--16.4 & 8.8 & 26.0 & \cellcolor{orange!15}--17.2 \\
GPT-4o & 97.1 & 3.0 & \cellcolor{blue!15}94.1 & 60.9 & 4.9 & \cellcolor{blue!15}56.0 \\
\hline
\end{tabular}

\vspace{1em}

\begin{tabular}{l@{\hspace{2pt}}ccc@{\hspace{4pt}}ccc}
\hline
& \multicolumn{3}{c}{\textbf{Antipresupp.}} & \multicolumn{3}{c}{\textbf{Deduct. Reason.}} \\
\cmidrule(lr){2-4} \cmidrule(lr){5-7}
\textbf{Model} & \textbf{l=1} & \textbf{l=0} & \textbf{$\Delta_{\text{cond}}$} & \textbf{l=1} & \textbf{l=0} & \textbf{$\Delta_{\text{cond}}$} \\
\hline
Mistral-7B & 58.8 & 88.9 & \cellcolor{orange!15}--30.0 & 24.0 & 78.8 & \cellcolor{orange!15}--54.8 \\
Mixtral-8x7B & 77.8 & --- & --- & 39.8 & 78.0 & \cellcolor{orange!15}--38.3 \\
Olmo-2-7B & 41.2 & 51.2 & \cellcolor{orange!15}--10.1 & 6.3 & 52.9 & \cellcolor{orange!15}--46.6 \\
Olmo-2-13B & 27.6 & 100.0 & \cellcolor{orange!15}--72.4 & 3.1 & 0.9 & \cellcolor{gray!15}2.3 \\
Olmo-2-32B & 78.0 & 85.7 & \cellcolor{orange!15}--7.8 & 2.2 & 44.0 & \cellcolor{orange!15}--41.7 \\
LLaMA-8B & 41.3 & 47.1 & \cellcolor{orange!15}--5.8 & 4.4 & 1.1 & \cellcolor{gray!15}3.3 \\
Qwen-3-8B & 61.5 & --- & --- & --- & 0.0 & --- \\
Qwen-3-14B & 42.3 & 50.0 & \cellcolor{orange!15}--7.7 & 0.0 & 0.0 & \cellcolor{gray!15}0.0 \\
Phi-4-14B & 73.0 & 94.7 & \cellcolor{orange!15}--21.8 & 100.0 & 5.1 & \cellcolor{blue!15}94.9 \\
Prometheus-14B & 56.7 & 92.2 & \cellcolor{orange!15}--35.4 & 100.0 & 2.0 & \cellcolor{blue!15}98.0 \\
Claude-Sonnet-4.5 & 100.0 & 100.0 & \cellcolor{gray!15}0.0 & 88.9 & 29.2 & \cellcolor{blue!15}59.6 \\
GPT-4o & 64.4 & 100.0 & \cellcolor{orange!15}--35.6 & 91.7 & 13.5 & \cellcolor{blue!15}78.1 \\
GPT-4.1 & 97.2 & 100.0 & \cellcolor{gray!15}--2.8 & 97.3 & 49.5 & \cellcolor{blue!15}47.7 \\
GPT-5 & 100.0 & 100.0 & \cellcolor{gray!15}0.0 & 100.0 & --- & --- \\
\hline
\end{tabular}

\caption{\footnotesize
\textbf{Conditional item-level relationship between listener and speaker performance.}
For each model and task, we report the probability of a correct speaker response conditional on a correct listener judgment ($l{=}1$) and on an incorrect listener judgment ($l{=}0$), along with their difference ($\Delta_{\text{cond}}$). 
%($\Delta_{\text{cond}} = P(\text{task}\mid l{=}1) - P(\text{task}\mid l{=}0)$). 
Cell colors indicate effect magnitude: 
\textcolor{orange!80}{negative} ($\Delta < -5$),
\textcolor{gray}{negligible} ($-5 \leq \Delta < 5$),
and \textcolor{blue!80}{positive} ($\Delta \geq 5$).
Dashes (---) indicate undefined conditional probabilities due to the absence of items in the corresponding listener condition.}
%All values are percentages.}
% \caption{\footnotesize
% \textbf{Conditional item-level relationship between listener and speaker performance.}
% For each model and task (FP = False Presuppositions), we report the probability of a correct speaker response conditional on a correct listener judgment ($l{=}1$) and on an incorrect listener judgment ($l{=}0$), along with their difference ($\Delta_{\text{cond}}$). 
% %($\Delta_{\text{cond}} = P(\text{task}\mid l{=}1) - P(\text{task}\mid l{=}0)$). 
% Cell colors indicate effect magnitude: \textcolor{orange!80}{negative}
% ($\Delta < -5$), \textcolor{gray}{negligible} ($-5 \leq \Delta \leq 0$),
% \textcolor{cyan!80}{small positive} ($0 < \Delta < 20$), and
% \textcolor{blue!80}{strong positive} ($\Delta \geq 20$).
% Dashes (---) indicate undefined conditional probabilities due to the absence of items in the corresponding listener condition.}
% %All values are percentages.}
\label{tab:conditional_analysis_all_results}
\end{table}

While aggregate accuracy comparisons reveal robust speaker–listener asymmetries, they do not address whether listener competence supports speaker performance at the level of individual items.  
To examine this, we analyze the item-level conditional relationship between listening and speaking across the three tasks (Table~\ref{tab:conditional_analysis_all_results}). 
Positive $\Delta_{\text{cond}}$ values indicate that correct listener judgments predict successful speaker performance, whereas zero or negative values indicate little or no such relationship.
%Positive $\Delta_{\text{cond}}$ values indicate cases where correct listener judgments most strongly predict successful speaker performance, whereas zero or negative values indicate little or no predictive relationship between judging and generation.

Across tasks, the conditional analysis reveals that high listener accuracy does \textit{not} generally translate into higher speaker accuracy on the same items. Instead, the relationship between judging and generation varies by pragmatic phenomenon.

%%%% False Presuppositions 
For \textbf{False Presuppositions}, the conditional relationship differs sharply across models and datasets.
For GPT-4o, listener judgments are strongly predictive of successful
generation: items that are correctly judged are also very likely to be handled
correctly in generation, yielding large positive conditional effects
($\Delta_{\text{cond}} = 56$–94).
By contrast, for Mistral-7B they are only negligible or small positive, and for LLaMA-8B, they are even negative: items that are correctly judged are less likely to be handled correctly in generation.

%%%% Antipresuppositions
For \textbf{Antipresuppositions,} conditional effects are predominantly negative across models.
For most systems, including Mistral-7B, Prometheus-14B, and
GPT-4o, speaker accuracy is lower on items that are judged correctly than on items judged incorrectly.
Thus, even when models successfully recognize violations of Maximize Presupposition!, this knowledge does not facilitate -- and may even interfere with -- correct trigger generation.
Only a few models (GPT-4.1, GPT-5, Claude-Sonnet-4.5) show near-zero conditional effects, primarily due to ceiling performance in both roles.
%Overall, the antipresupposition results provide little evidence that pragmatic judgment supports pragmatic generation at the item level.

%%%%Deductive Reasoning.
For \textbf{Deductive Reasoning}, results are mixed.
Several models, especially proprietary ones (Phi-4-14B, Prometheus-14B, Claude-Sonnet-4.5, GPT-4o, GPT-4.1), exhibit strong positive conditional effects, indicating that correct logical judgments are associated with higher success in generating valid conclusions.
In contrast, most open-weight models show large negative effects, suggesting a misalignment between evaluating and generating deductive inferences.
For some models (Qwen-3-8B, Qwen-3-14B, Olmo-2-13B, Claude-Sonnet-4.5) estimates should be interpreted cautiously due to instruction-following failures (Table \ref{tab:parsing_rates}).

Overall, this conditional analysis indicates that pragmatic listening and speaking are often item-level independent.
Positive coupling between judging and generation emerges mainly for False Presuppositions and Deductive Reasoning in large proprietary models, but is largely absent for many open-weight systems and even entirely absent for most models in Antipresuppositions. 
% This reinforces the view that pragmatic listening and speaking are partially distinct capacities in current LLMs, even when tested on the same underlying items.
This reinforces the view that pragmatic listening and speaking are not reliably coupled capacities in current LLMs: while some models show partial alignment in specific tasks, others exhibit clear dissociation, even when evaluated on the same underlying items.

%%%%%%%%%%%%%%%%%%%%%%%%%%%%%%%%%%%%%
%%%%%%%%%%%%%%%%%%%%%%%%%%%%%%%%%%%%%%
\section{Discussion and Conclusion}

In this paper, we examined whether LLMs' performance as pragmatic \emph{listeners}, i.e., judging the appropriateness of linguistic outputs, aligns with their performance as pragmatic \emph{speakers}, i.e., generating pragmatically appropriate language. 

Across three pragmatic tasks -- Antipresuppositions, False Presuppositions, and Deductive Reasoning -- we observed a consistent asymmetry between the models' performance as pragmatic speakers and as pragmatic listeners. Many models achieved substantially higher accuracy when judging pragmatic appropriateness than when generating pragmatically appropriate outputs themselves. 
Notably, this pattern persisted even when speaker performance required only minimal generation (e.g., single-word responses in Antipresuppositions and Deductive Reasoning).
The asymmetry was most pronounced for the open-weight and mid-sized models, which often failed on the pragmatic speaker task while performing substantially better as pragmatic listeners. For larger and proprietary models, speaker and listener performance was more closely aligned, but still not identical: even models with strong generation abilities exhibited non-trivial mismatches between solving pragmatic tasks and evaluating their solutions. 
Crucially, item-level analyses showed that this asymmetry is not merely due to overall difficulty or scaling: across tasks, correct listener judgments did not reliably predict successful speaker behavior on the same underlying items. 
In some settings -- most notably Antipresuppositions -- correct judgments were even associated with lower generation success, indicating that recognizing pragmatic violations does not necessarily support correct generation.

Overall, these results echo long-standing observations from psycholinguistics that language comprehension and production are closely related yet not identical tasks, and that success in one does not straightforwardly entail success in the other. %Our findings suggest that LLM-based pragmatic competence exhibits a similar pattern: the ability to recognize pragmatic adequacy does not guarantee the ability to generate pragmatically appropriate language.
While we do not claim analogous mechanisms, our findings suggest that LLM behavior exhibits a comparable asymmetry: recognizing pragmatic adequacy does not reliably translate into generating pragmatically appropriate language.
Moreover, related divergences between judging and generation have also been reported outside the domain of pragmatics \cite{piot2025llmsevaluateannotaterevisiting}, suggesting that this pattern may reflect a more general property of current LLM behavior rather than task-specific artifacts.

\paragraph{Implications.}
The observed asymmetry between pragmatic listening and speaking has implications for using LLMs both as generators and evaluators.

First, the prevalence of instruction-following failures raises concerns for the use of LLMs as automatic judges. Several models produced outputs that violated explicit format constraints and could not be reliably parsed (Table~\ref{tab:parsing_rates}), limiting  the interpretability of their evaluation performance. %and highlights that apparent judgment accuracy may depend as much on instruction adherence as on underlying task competence.
Moreover, evaluation reliability was not a stable property of a given model, but varied across tasks and prompt types, underscoring the need to validate LLM-as-a-judge setups on a per-task basis.

Second, strong performance in pragmatic listener tasks should not be taken as evidence that a model possesses equally robust pragmatic generation abilities. %Even when models accurately identify pragmatic violations or appropriate responses, this competence often fails to translate into correct behavior in generation. 
The observed misalignment between the two roles aligns with recent work questioning the use of metalinguistic prompting as a probe of model competence.
For example, when probing syntactic competence, \citet{hu-levy-2023-prompting} show that metalinguistic judgments elicited via prompting can diverge substantially from quantities derived directly from models’ representations, and that poor prompting performance does not necessarily reflect missing linguistic knowledge. 
Our results extend this critique to pragmatic evaluation: strong performance in listener-style judgment tasks does not reliably reflect a model's ability to generate pragmatically appropriate responses. Together, these findings suggest that prompt-based evaluation tasks capture only task-specific behaviors rather than providing a transparent indicator of a model's broader communicative competence.

With this, our findings raise questions about current evaluation practices. Many benchmarks targeting pragmatic abilities rely predominantly on listener-style tasks \cite{sileo-etal-2022-pragmatics, sravanthi-etal-2024-pub, ma-etal-2025-pragmatics}, implicitly assuming that success in evaluation reflects underlying generative competence. Our findings challenge this assumption and point toward the need for evaluation frameworks that assess language models as integrated systems, jointly considering generation and evaluation rather than treating them as interchangeable proxies for one another.

\section*{Limitations}

This study has several limitations that should be taken into account.

First, our analysis is restricted to three pragmatic phenomena: False Presuppositions, Antipresuppositions, and Deductive Reasoning. While these tasks capture distinct aspects of pragmatic behavior, they do not cover the full range of pragmatic phenomena. The observed speaker–listener differences may therefore not generalize to other pragmatic domains, such as implicature or conversational repair.

Second, our listener evaluations rely on a single prompt formulation per task, where possible, closely modeled on the original task or annotation instructions. While this choice ensures comparability with prior work, alternative prompt designs or evaluative framings may yield different results. 

Third, the experiments span multiple languages: two tasks are conducted in German, while Deductive Reasoning is evaluated in English.
Although this reflects the languages in which datasets for these phenomena are currently available, it limits our ability to disentangle pragmatic effects from potential language-specific influences. 
Future work could systematically vary language by testing the same pragmatic tasks across languages to assess the robustness of speaker--listener differences.

Fourth, our analysis is purely behavioral. We do not examine model internals or training data, and therefore make no claims about the mechanisms underlying the observed differences. While our results demonstrate systematic differences between pragmatic evaluation and pragmatic generation, explaining why these differences arise remains an interesting open question for future work.

Finally, all experiments rely on a fixed inference setup: we use deterministic decoding for open-weight models to ensure reproducibility and stable item-level comparisons, and evaluate all models in a zero-shot setting (cf. Appendix~\ref{sec:appendix_implement_details}). While these choices follow common practice in in similar evaluation setups, alternative decoding strategies, in-context learning, or task-specific fine-tuning may influence model behavior and yield different absolute performance. Exploring such interventions is left to future work.

\section*{Ethical considerations}
%\judith{openreview: Did you discuss any potential risks of your work?}

This work examines pragmatic evaluation and generation in large language models using existing benchmark-style tasks. We do not identify any direct ethical risks arising from our methodology or findings.
All datasets used in this study are publicly available and were originally collected for research purposes, and our use of these resources is consistent with their intended research use. The experiments involve no human subjects, personal data, or sensitive attributes beyond those already present in the original datasets. We do not introduce new human annotations, nor do we deploy the models in real-world decision-making settings.

\section*{Acknowledgements}
We acknowledge financial support from the
project “SAIL: SustAInable Life-cycle of Intelligent Socio-Technical Systems" (Grant ID NW21-059A), an initiative of the Ministry of Culture and Science of the State of North Rhine-Westphalia.

\bibliography{custom}

\appendix

\section{Appendix}
\label{sec:appendix}

\subsection{Implementation Details}
\label{sec:appendix_implement_details}

\paragraph{Models and Inference Setup.}

We evaluate a diverse set of large language models, including both open-weight and proprietary systems, as described in Section~\ref{sec:experiment}.
Concretely, we used the following models for our experiments:
\vspace{0.4em}
\begin{itemize}%[noitemsep, topsep=0pt]
    \item \href{https://huggingface.co/mistralai/Mistral-7B-Instruct-v0.2}{Mistral-7B-Instruct-v0.2}
    \item \href{https://huggingface.co/mistralai/Mixtral-8x7B-Instruct-v0.1}{Mixtral-8x7B-Instruct-v0.1} 
    \item \href{https://huggingface.co/allenai/OLMo-2-1124-7B-Instruct}{OLMo-2-1124-7B-Instruct} 
    \item \href{https://huggingface.co/allenai/OLMo-2-1124-13B-Instruct}{OLMo-2-1124-13B-Instruct} 
    \item \href{https://huggingface.co/allenai/OLMo-2-1124-13B-Instruct}{OLMo-2-0325-32B-Instruct} 
    \item \href{https://huggingface.co/meta-llama/Llama-3.1-8B-Instruct}{Llama-3.1-8B-Instruct} 
    \item \href{https://huggingface.co/Qwen/Qwen3-8B}{Qwen3-8B} 
    \item \href{https://huggingface.co/Qwen/Qwen3-14B}{Qwen3-14B} 
    \item \href{https://huggingface.co/microsoft/phi-4}{Phi-4} 
    \item \href{https://huggingface.co/Unbabel/M-Prometheus-14B}{M-Prometheus-14B} 
    \item \href{https://www.anthropic.com/claude/sonnet}{Claude-Sonnet-4.5} 
    \item \href{https://openai.com/de-DE/index/hello-gpt-4o}{GPT-4o} 
    \item \href{https://platform.openai.com/docs/models/gpt-4.1}{GPT-4.1} 
    \item \href{https://platform.openai.com/docs/models/gpt-5}{GPT-5} 
\end{itemize}
\vspace{0.4em}
All models were queried in a zero-shot setting using fixed prompts; no task-specific fine-tuning or hyperparameter search was performed.

For open-weight models, we disabled sampling and used deterministic decoding with a maximum generation length sufficient to cover the expected output formats for each task.
This choice reduces stochastic variability and ensures reproducibility, which is particularly important for stable item-level comparisons, and it follows prior work using similar evaluation setups (e.g., \citealp{Bavaresco2024-LLMsasJudge}). No additional decoding constraints (e.g., nucleus sampling or repetition penalties) were applied. Proprietary models were accessed via their respective APIs.

\paragraph{Implementation and Compute.}
All experiments with open-weight models were implemented in Python~3.9 using PyTorch and the Hugging Face transformers library. Experiments were run on a single NVIDIA RTX A6000 GPU with CUDA acceleration. Depending on model size and task, generating all responses for a model required between approximately 1 and 48 GPU hours.
Proprietary models were accessed via their respective APIs. Total API costs were modest (in the range of a few USD per provider).

\paragraph{Output Parsing and Normalization.}
Across all tasks, models were instructed to respond in a strictly constrained output format (e.g., a single word or label) (Figure \ref{fig:all_example_prompts}).
Nevertheless, some models produced outputs that mixed target labels with free-form text, explanations, or formatting deviations.
To ensure comparability across models, we applied a rule-based parser that maps model outputs to the expected label set where possible.

\paragraph{Parsing Statistics.}
Adherence to instructions in the prompts varied substantially across models, tasks, and roles.
Table~\ref{tab:parsing_rates} reports parsing rates for all model--task--role combinations.

Unparsable outputs primarily resulted from violations of explicit output constraints (e.g., providing explanations despite instructions to respond with a single word or label, or using formats outside the specified option set).
Instruction-following failures are most pronounced in generation-based speaker tasks, but their severity differs across pragmatic settings.
For False Presuppositions, where we modeled the listener task, parsing reliability is uniformly high across models.
For Antipresuppositions, parsing reliability is likewise generally high, with Olmo-2-13B as the main exception (38.1\% parsable speaker outputs).
For Deductive Reasoning, parsing failures are substantially more severe. For instance, Qwen3-14B produced no parsable outputs in the speaker condition (0\% parsing), and Qwen3-8B failed to produce parsable outputs in both the speaker and listener conditions. Olmo-2-13B achieved near-zero parsing coverage in the speaker condition (5\%), despite perfect coverage in the listener condition.
In addition, LLaMA-8B, Qwen3-32B, and Claude-Sonnet-4.5, show only moderate parsing reliability in the Deductive Reasoning speaker task (around 50\%). Notably, Claude-Sonnet-4.5 also exhibits extremely low parsing reliability in the Deductive Reasoning listener condition (5\%), markedly lower than all other evaluated models.
We therefore report results for these models and tasks, but interpret their performance with caution.

\subsection{Scientific Artifacts}
\label{app:scientific-artifacts}

We used publicly available datasets, experimental materials, and model implementations, all in accordance with their intended research use and licensing terms. 
The datasets and prompts consist of constructed linguistic examples and model-generated outputs and do not contain personal data or information that identifies individual people; no anonymization was therefore required.
Artifacts created in this work (including prompts and evaluation scripts) are intended exclusively for research and reproducibility purposes and are compatible with the access conditions of the original data sources.

\subsection{Use of AI Assistants}
AI assistants were used during manuscript preparation only for limited linguistic editing to improve clarity and style, and for writing auxiliary code (e.g., for visualizations). They were not used for scientific reasoning, evaluation decisions, or interpretation of results; all analyses and conclusions were drawn by the authors.%\newpage

\section{Additional Results}\label{sec:appendix_additional_results}

\begin{figure*}[!tbp]
    \centering
    \includegraphics[width=\textwidth]{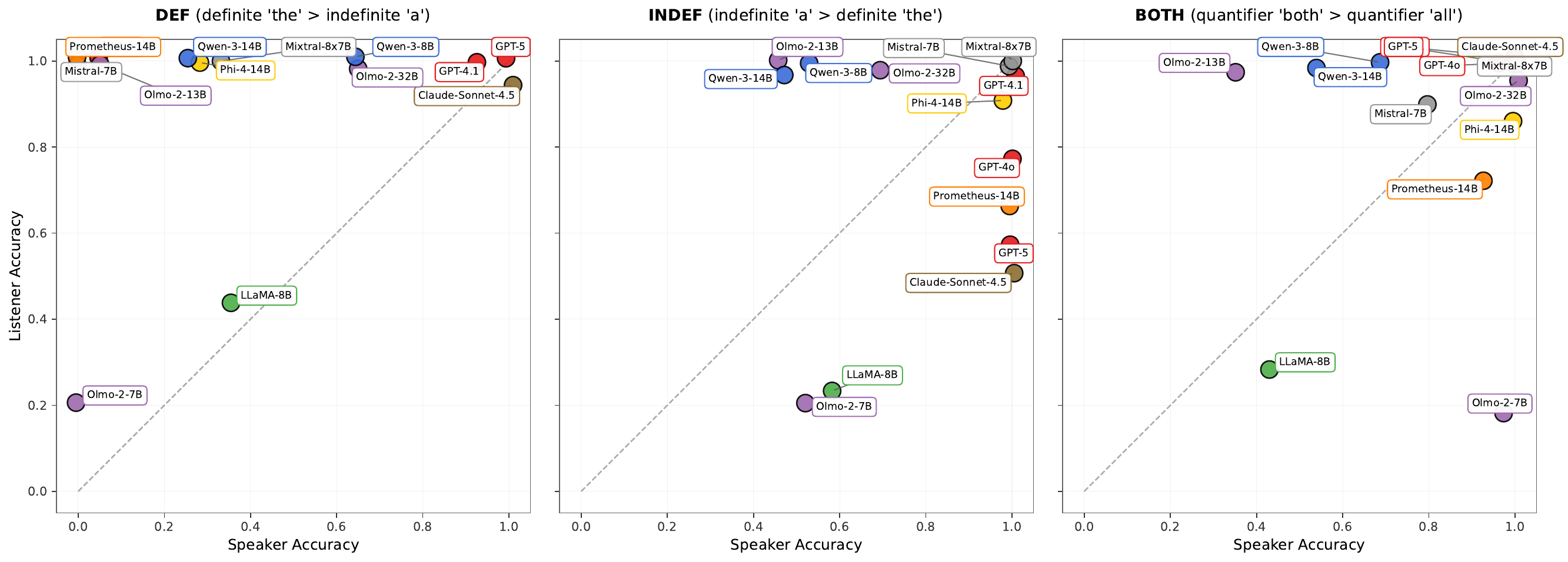}
    \caption{\footnotesize\textbf{Speaker vs.\ Listener accuracy split by condition for Antipresuppositions task}. 
    \textit{DEF} = MP! demands definite determiner, \textit{INDEF} = MP! demands indefinite determiner, \textit{BOTH} = MP! demands the quantifier 'both'. 
    Each point is one model; colors indicate model families. 
    The diagonal marks equal speaker and listener accuracy, so points above the line correspond to models that are better in the listener task than in the speaker task.}
    \label{fig:antipsp_family_task_judge}
\end{figure*}

%%%%%%%%%%%%%%%%%%%%%%%%%%%%%%%%%%%%%%%%%%%%%%%%%%%%%%%%%%%%%%%%%%%%
%%%%%%%%%%%%%%%%% Antipresuppositions
\paragraph{Contrast-wise analysis for Antipresuppositions.}
Figure~\ref{fig:antipsp_family_task_judge} breaks down pragmatic speaker and listener performance by presuppositional contrast, revealing that the overall speaker–listener asymmetry is not uniform across conditions but depends on the type of presupposition trigger involved.

In the \texttt{DEF} condition (i.e., where the MP! principle requires the definite determiner "the"), most models perform substantially better as pragmatic listeners than as speakers.
Speaker accuracy is often at or below chance, while listener accuracy approaches ceiling. This indicates that models reliably recognize violations of MP! when evaluating completed sentences, despite failing to consistently select the definite form during generation.

By contrast, in the \texttt{INDEF} condition (i.e., where the MP! principle requires the indefinite determiner "a"), the asymmetry largely disappears. Many models achieve near-ceiling speaker accuracy when generating indefinite
forms, reflecting a strong default generation preference for weaker
presupposition triggers. In this condition, listener accuracy is sometimes lower than speaker accuracy, suggesting that recognizing the pragmatic infelicity of an unnecessary definite is harder than producing the correct indefinite determiner.
This pattern aligns with earlier findings by \citet{sieker-zarriess-2023-language}, who show that masked language models (BERT and variants) strongly favor indefinite determiners across conditions. They argue that this bias may partly reflect models’ tendency to reproduce surface patterns present in the prompt, where an indefinite determiner is used in the context sentence.

The \texttt{BOTH} condition (i.e., where the MP! principle requires the quantifier "both") exhibits a more heterogeneous pattern. Some models show higher listener than speaker accuracy, while others perform better as speakers, indicating that this contrast interacts more variably with model-specific generation and evaluation strategies.

Taken together, these results show that the speaker–listener asymmetry in Antipresuppositions arises most clearly when pragmatic reasoning conflicts with models’ default generation preferences. When the pragmatically appropriate form is also the model’s preferred continuation (as in the \texttt{INDEF} condition), speaker performance is high and the asymmetry is reduced. When pragmatic constraints require overriding this bias (as in the \texttt{DEF} condition), models struggle in generation while remaining highly sensitive to infelicity in evaluation.

\vspace{2em} 
%%%%%%%%%%%%%%%%%%%%%%%%%%%%%%%%%%%%%%%%%%%%%%%%%
%%%%%%%%%%%%%%%%%%%%%%%%%%%%%%%%%%%%%%%%%%%%%%%%%

%%%%%%% Tabellen zu task vs. judge accuracies
%%%%%%% False Presuppositions 
\newblock
\begin{table}[!htbp]
\centering
\small
\setlength{\tabcolsep}{4pt}
% ====================
% SCENARIOS
% ====================
\begin{tabular}{lrrr}
\hline
\multicolumn{4}{c}{\textbf{False Presuppositions — Scenarios}} \\
\hline
\textbf{Model} & \textbf{Listener Acc.} & \textbf{Speaker Acc.} & \textbf{$\Delta$} \\
\hline
Mistral-7B & 0.33 & 0.02 & \cellcolor{blue!15}{\textbf{+0.31}} \\
LLaMA-8B   & 0.44 & 0.16 & \cellcolor{blue!15}{+0.28} \\
GPT-4o     & 0.78 & 0.84 & \cellcolor{orange!15}{\textbf{--0.06}} \\
\hline
\end{tabular}

\vspace{1em}
% ====================
% CLAIMS
% ====================
\begin{tabular}{lrrr}
\hline
\multicolumn{4}{c}{\textbf{False Presuppositions — Claims}} \\
\hline
\textbf{Model} & \textbf{Listener Acc.} & \textbf{Speaker Acc.} & \textbf{$\Delta$} \\
\hline
Mistral-7B & 0.38 & 0.10 & \cellcolor{blue!15}{+0.28} \\
LLaMA-8B   & 0.43 & 0.21 & \cellcolor{blue!15}{+0.22} \\
GPT-4o     & 0.63 & 0.38 & \cellcolor{blue!15}{+0.25} \\
\hline
\end{tabular}

\caption{\footnotesize
\textbf{False Presuppositions:} Listener (judge) vs.\ speaker (generation) accuracy.
$\Delta =$ Listener -- Speaker.
\textcolor{blue!60}{Purple} cells indicate higher listener than speaker accuracy; \textcolor{orange!80}{orange} cells indicate the reverse.
The largest positive and negative $\Delta$ values are highlighted in bold. 
}
\label{tab:fpsp_task_vs_judge_combined}
\end{table}

\vspace{2em}   

%%%%%%% Antipresuppositions 
\begin{table}[!htbp]
\centering
\small
\setlength{\tabcolsep}{4pt}
\begin{tabular}{lrrr}
\hline
\textbf{Model} & \textbf{Listener Acc.} & \textbf{Speaker Acc.} & \textbf{$\Delta$} \\
\hline
Mistral-7B          & 0.96 & 0.60 & \cellcolor{blue!15}+0.36 \\
Mixtral-8x7B        & 1.00 & 0.78 & \cellcolor{blue!15}+0.22 \\
Olmo-2-7B           & 0.20 & 0.49 & \cellcolor{orange!15}\textbf{--0.29} \\
Olmo-2-13B          & 0.99 & 0.28 & \cellcolor{blue!15}\textbf{+0.71} \\
Olmo-2-32B          & 0.97 & 0.78 & \cellcolor{blue!15}+0.19 \\
LLaMA-8B            & 0.32 & 0.45 & \cellcolor{orange!15}--0.13 \\
Qwen-3-8B           & 1.00 & 0.62 & \cellcolor{blue!15}+0.39 \\
Qwen-3-14B          & 0.98 & 0.42 & \cellcolor{blue!15}+0.56 \\
Phi-4-14B           & 0.92 & 0.75 & \cellcolor{blue!15}+0.18 \\
Prometheus-14B      & 0.80 & 0.64 & \cellcolor{blue!15}+0.16 \\
Claude-Sonnet-4.5   & 0.82 & 1.00 & \cellcolor{orange!15}--0.18 \\
GPT-4o              & 0.92 & 0.67 & \cellcolor{blue!15}+0.25 \\
GPT-4.1             & 0.99 & 0.97 & \cellcolor{gray!15}+0.02 \\
GPT-5               & 0.86 & 1.00 & \cellcolor{orange!15}--0.14 \\
\hline
\end{tabular}
\caption{\footnotesize
\textbf{Antipresuppositions:} Listener (judge) vs.\ speaker (generation) accuracy.
$\Delta =$ Listener -- Speaker.
\textcolor{blue!60}{Purple} cells indicate higher listener than speaker accuracy;
\textcolor{orange!80}{orange} cells indicate the reverse.
The largest positive and negative $\Delta$ values are highlighted in bold. 
}
\label{tab:antipsp_task_vs_judge}
\end{table}

%%%%%%%%%% Deductive Reasoning
\begin{table}[!htbp]
\centering
\small
\setlength{\tabcolsep}{4pt}
\begin{tabular}{lrrr}
\hline
\textbf{Model} & \textbf{Listener Acc.} & \textbf{Speaker Acc.} & \textbf{$\Delta$} \\
\hline
Mistral-7B          & 0.56 & 0.48 & \cellcolor{blue!15}+0.08 \\
Mixtral-8x7B        & 0.54 & 0.57 & \cellcolor{orange!15}--0.03 \\
Olmo-2-7B           & 0.53 & 0.28 & \cellcolor{blue!15}+0.25 \\
Olmo-2-13B          & 0.36 & 0.02 & \cellcolor{blue!15}+0.34 \\
Olmo-2-32B          & 0.49 & 0.23 & \cellcolor{blue!15}+0.26 \\
LLaMA-8B            & 0.51 & 0.03 & \cellcolor{blue!15}+0.48 \\
Qwen-3-8B           & 0.00 & 0.00 & \cellcolor{gray!15}+0.00 \\
Qwen-3-14B          & 0.54 & 0.00 & \cellcolor{blue!15}\textbf{+0.54} \\
Phi-4-14B           & 0.45 & 0.48 & \cellcolor{orange!15}--0.03 \\
Prometheus-14B      & 0.43 & 0.44 & \cellcolor{orange!15}--0.01 \\
Claude-Sonnet-4.5   & 0.05 & 0.32 & \cellcolor{orange!15}--0.27 \\
GPT-4o              & 0.47 & 0.50 & \cellcolor{orange!15}--0.03 \\
GPT-4.1             & 0.41 & 0.69 & \cellcolor{orange!15}\textbf{--0.28} \\
GPT-5               & 1.00 & 1.00 & \cellcolor{gray!15}+0.00 \\
\hline
\end{tabular}

\caption{\footnotesize
\textbf{Deductive Reasoning:} Listener (judge) vs.\ speaker (generation) accuracy.
$\Delta =$ Listener -- Speaker.
\textcolor{blue!60}{Purple} cells indicate higher listener than speaker accuracy;
\textcolor{orange!80}{orange} cells indicate the reverse.
}
\label{tab:deductreason_task_vs_judge}
\end{table}

%%%%%%%%%%%%%%%%%%%%%%%%%%%%%
%%%%%% Parsing rates (extended with Deductive Reasoning)
\begin{table*}[!htbp]
\centering
\small
\resizebox{\textwidth}{!}{%
\begin{tabular}{lrrrrr}
\hline
\textbf{Model} 
& \textbf{FP Listener (\%)} 
& \textbf{AntiPSP Listener (\%)} 
& \textbf{AntiPSP Speaker (\%)} 
& \textbf{Deduct. Reason Listener (\%)} 
& \textbf{Deduct. Reason Speaker (\%)} \\
\hline
Mistral-7B            & 100.0 & 100.0 & 93.3  & 89.4 & 100.0 \\
Mixtral-8x7B          & 100.0 & 100.0 & 100.0 & 100.0 & 98.9 \\
Olmo-2-7B             & 100.0 & 98.8  & 65.5  & 99.4 & 92.2 \\
Olmo-2-13B            & 100.0 & 87.1  & 38.1  & 5.6  & 100.0 \\
Olmo-2-32B            & 99.8  & 100.0 & 100.0 & 56.1 & 100.0 \\
LLaMA-8B              & 91.1  & 73.6  & 98.4  & 49.4 & 92.8 \\
Qwen-3-8B             & 98.8  & 88.9  & 99.6  & 0.0  & 0.0 \\
Qwen-3-14B            & 100.0 & 81.8  & 65.9  & 0.0  & 85.0 \\
Phi-4-14B             & 100.0 & 100.0 & 99.2  & 100.0 & 100.0 \\
Prometheus-14B        & 100.0 & 100.0 & 97.6  & 100.0 & 100.0 \\
Claude-Sonnet-4.5     & 100.0 & 100.0 & 100.0 & 55.0 & 5.0 \\
GPT-4o                & 100.0 & 100.0 & 98.8  & 100.0 & 100.0 \\
GPT-4.1               & 100.0 & 100.0 & 100.0 & 100.0 & 100.0 \\
GPT-5                 & 100.0 & 100.0 & 100.0 & 100.0 & 100.0 \\
\hline
\end{tabular}
}
\caption{\footnotesize
\textbf{Instruction-following reliability across tasks.}
Percentages indicate the proportion of outputs that could be parsed into the
target evaluation format for each task
(FP = False Presupposition listener task;
AntiPSP = Antipresuppositions listener and speaker tasks;
Deductive Reasoning = listener and speaker tasks).
}
\label{tab:parsing_rates}
\end{table*}

%%%%%%% Antipresuppositions task-judge scatterplot by condition

\end{document}